\documentclass[conference]{IEEEtran}
\IEEEoverridecommandlockouts
\usepackage{cite}
\usepackage{subfig}
\usepackage{textcase}
\usepackage{caption}
\usepackage[tablename=TABLE]{caption}
\DeclareCaptionTextFormat{up}{\MakeTextLowercase{#1}}
\usepackage{siunitx}
\usepackage{graphicx} 
\usepackage{amsmath}
\usepackage{amssymb}
\usepackage{amsmath,amssymb,amsfonts}
\usepackage{algorithmic}
\usepackage{graphicx}
\usepackage{textcomp}
\usepackage{xcolor}
\def\BibTeX{{\rm B\kern-.05em{\sc i\kern-.025em b}\kern-.08em
    T\kern-.1667em\lower.7ex\hbox{E}\kern-.125emX}}
\begin{document}

\title{EGFR Mutation Prediction of Lung Biopsy Images using Deep Learning}
\makeatletter
\newcommand{\linebreakand}{%
  \end{@IEEEauthorhalign}
  \hfill\mbox{}\par
  \mbox{}\hfill\begin{@IEEEauthorhalign}
}
\makeatother
\author{\IEEEauthorblockN{Ravi Kant Gupta}
\IEEEauthorblockA{\textit{Department of Electrical Engineering} \\
\textit{Indian Institute of Technology, Bombay}\\
Mumbai, India \\
184070025@iitb.ac.in}
\and
\IEEEauthorblockN{Shivani Nandgaonkar}
\IEEEauthorblockA{\textit{Department of Electrical Engineering} \\
\textit{Indian Institute of Technology, Bombay}\\
Mumbai, India \\
17d070003@iitb.ac.in}
\and
\IEEEauthorblockN{Nikhil Cherian Kurian}
\IEEEauthorblockA{\textit{Department of Electrical Engineering} \\
\textit{Indian Institute of Technology, Bombay}\\
Mumbai, India \\
nikhilkurian@iitb.ac.in}
\linebreakand
\IEEEauthorblockN{Tripti Bameta}
\IEEEauthorblockA{\textit{Computational Pathology Laboratory} \\
\textit{Tata Memorial Centre-ACTREC, HBNI}\\
Mumbai, India \\
tripti.bameta@gmail.com}
\and
\IEEEauthorblockN{Subhash Yadav}
\IEEEauthorblockA{\textit{Department of Pathology} \\
\textit{Tata Memorial Centre-ACTREC, HBNI}\\
Mumbai, India \\
drsubhashyadav23@gmail.com}
\and
\IEEEauthorblockN{Rajiv Kumar Kaushal}
\IEEEauthorblockA{\textit{Department of Pathology} \\
\textit{Tata Memorial Centre-TMH, HBNI}\\
Mumbai, India \\
rajiv.kaushal@gmail.com}
\linebreakand
\IEEEauthorblockN{Swapnil Rane}
\IEEEauthorblockA{\textit{Department of Pathology} \\
\textit{Tata Memorial Centre-ACTREC, HBNI}\\
Navi mumbai, India \\
raneswapnil82@gmail.com}
\and
\IEEEauthorblockN{Amit Sethi}
\IEEEauthorblockA{\textit{Department of Electrical Engineering} \\
\textit{Indian Institute of Technology, Bombay}\\
Mumbai, India \\
asethi@iitb.ac.in}

}

\maketitle

\begin{abstract}
% The primary diagnosis of lung cancer, which is a leading cause of death worldwide, includes cancer detection, histological subtyping, and detection of key driver mutations, especially the EGFR mutation. These tasks require visual inspection of lung biopsies by experienced pathologists. To reduce the time and cost of these tasks we trained deep learning models on whole slide images of The Cancer Genome Atlas (TCGA) from the US and a private dataset from India. With the use of weakly supervised learning and extensive ablation studies, we demonstrated the effectiveness of our approach. We achieved an average area under the curve (AUC) of 0.964 for tumor detection,  and 0.942 for histological subtyping between adenocarcinoma and squamous cell carcinoma on the TCGA dataset. For EGFR detection, we achieved an average AUC of 0.864 on the TCGA dataset and 0.783 on the dataset from India. Our key learning points include the following. Firstly, there is no particular advantage of using a feature extractor layers trained on histology, if one is going to fine-tune the feature extractor on the target dataset. Secondly, selecting patches with high cellularity, presumably capturing tumor regions, is not always helpful, as the sign of a disease class may be present in the tumor-adjacent stroma.
The standard diagnostic procedure for targeted therapies in lung cancer treatment involve cancer detection, histological subtyping, and subsequent detection of key driver mutations, such as epidermal growth factor receptor (EGFR). Even though molecular profiling can uncover the driver mutation, the process is expensive and time-consuming. Deep learning-based image analysis offers a more economical alternative for discovering driver mutations directly from whole slide images (WSIs) of tissue samples stained using hematoxylin and eosin (H\&E). In this work, we used customized deep learning pipelines with weak supervision to identify the morphological correlates of EGFR mutation from hematoxylin and eosin-stained WSIs, in addition to detecting tumor and histologically subtyping it. We demonstrate the effectiveness of our pipeline by conducting rigorous experiments and ablation studies on two lung cancer datasets -- the cancer genome atlas (TCGA) and a private dataset from India. With our pipeline, we achieved an average area under the curve (AUC) of 0.964 for tumor detection and 0.942 for histological subtyping between adenocarcinoma and squamous cell carcinoma on the TCGA dataset. For EGFR detection, we achieved an average AUC of 0.864 on the TCGA dataset and 0.783 on the dataset from India. Our key findings are the following. Firstly, there is no particular advantage of using feature extractor layers trained on histology if there are differences in magnification. Secondly, selecting patches with high cellularity, presumably capturing tumor regions, is not always helpful, as the sign of a disease class may be present in the tumor-adjacent stroma. And finally, color normalization is still an alternative worth trying when compared to color jitter, even though their origins lie in opposing approaches to dealing with stain color variation. 
\end{abstract}


\section{Introduction}
Lung cancer is a leading cause of death worldwide \cite{b20}. Non-small cell lung cancer (NSCLC) and small-cell lung cancer (SCLC) are two major types of lung cancer of which the former is more common. NSCLC usually arises in a outer region of the lung and may look like pneumonia on chest X-ray~\cite{b21}. NSCLC has two major histologic variants -- lung adenocarcinoma (LUAD) and lung squamous cell carcinoma (LUSC) -- and one less prevalent subtype -- large cell carcinoma. Identification of lung cancer subtype is a key diagnostic step, because the major lung cancer subtypes -- LUAD and LUSC -- have different treatment regimens. Treatments also differ by major driver mutations, including epidermal growth factor receptor (EGFR) mutations that are present in about 20\% of LUAD, and anaplastic lymphoma receptor tyrosine kinase (ALK) rearrangements that are present in less than 5\% of LUAD~\cite{b19}. Lung biopsies stained using hematoxylin and eosin (H\&E), which are inexpensive and available in most pathology labs, are used to determine lung cancer subtype and stage. For EGFR mutation detection, the EGFR immunohistochemical (IHC) stain is not very reliable, and the molecular test is expensive, time consuming, and not widely available.

By solving image classification and prediction tasks, deep learning is poised to revolutionize the analyses of medical images, even though working with whole slide images (WSIs) is very challenging. Deep learning-based computational pathology methods require either manually annotated WSIs for full supervision or large datasets with slide-level labels for weak supervision. Slide-level labels may correspond to only small regions of a large gigapixel image. Consequently, numerous methods depend on annotation at pixel-level, patch-level, or region of interest (ROI)-level. These large WSIs are oftentimes challenging to inspect manually, which makes accurate interpretation a tedious task. Not all mutations are equally easy to detect in H\&E stained pathology images. Published mutation detection accuracies in held-out cases range from 1.000 for BRAF mutation in thyroid cancer \cite{b24} to 0.632 for NF1 mutation in Lung cancer \cite{b2}. Detecting EGFR mutation in lung cancer has been done with 0.826 AUC in whole slide images (WSIs) of formalin fixed paraffin embedded tissue \cite{b2}. In this study, we share how we pushed that accuracy higher, and what still needs to be done before such pipelines can be used in clinical settings, even if only for triaging. For instance, the distinction between LUAD and LUSC is not always clear, especially in scenarios where tumors are poorly differentiated. To predict the gene mutation manually is even more difficult and inconsistent even among experienced pathologists, because there are no reliable morphological signs to identify these mutations in H\&E. 

To handle aforementioned challenges, we propose a deep learning pipeline combined with weakly supervised learning. This pipeline is a step towards ushering inexpensive and timely tumor detection, histological subtyping, and mutation identification. We present our findings from developing convolution neural networks (CNNs) for such tasks based on TCGA WSIs from the US as well as an Indian dataset from the Tata Memorial Centre (TMC) in Mumbai. The work on TMC data was approved by the TMC Institutional Ethics Committee. These datasets contain gigapixel sized WSIs of formalin-fixed paraffin embedded (FFPE) tissue sections that are stained using inexpensive and ubiquitous H\&E stains and scanned at $40\times$ magnification. Figure\ref{slide} shows some snapshots of slides from both datasets. 

Our pipeline is based on several practical considerations. Firstly, to handle large size of WSIs, we segmented the tissue region of each WSI. Then we extracted patches from the segmented tissue of each WSI. We found a huge colors variations among extracted patches. This variation is due to staining protocols, habit of technician, reagent brands and color response of scanner. To mitigate this variation, we performed color normalization on extracted patches as shown in Figure \ref{fig2}. All the color normalised tissue patches of each WSI served as input to the CNN to create a set of low-dimensional feature embedding (Figure \ref{fig3}). Low-dimensional feature spaces are more suitable for faster training and reduced computational cost. By projecting patches to a low-dimensional space, the volume of data is reduced nearly 200 times and led to subsequent reduction of computational requirements to train deep learning models. These embeddings were used for all three tasks -- tumor detection, histological subtyping, and EGFR mutation prediction. Since visual features associated with the mutation cannot be reliably annotated by a pathologist, a weakly supervised learning technique was used for the third task. At a high-level during training and inference both, the model examines and ranks all patches in the tissue regions of WSI for their relevance in EGFR mutation prediction. An attention score is assigned to each patch on the basis of importance or contribution to the collective slide-level representation for a particular class. This attention score is used for the slide-level aggregation based on attention-based pooling, which computes the slide-level representation as the average of all patches weighted by their respective attention score, as shown in Figure \ref{fig4}.

We obtained high accuracy for the first two tasks in line with a previous study~\cite{b2}. For EGFR mutation, the manual data filtering of TCGA data in the previous studies is not clear, and yet we were able to obtain a higher AUC of 0.864. To do so, we performed extensive ablations studies that are listed in Section~\ref{sec:expt}. Among our key findings, we found that color normalization is still an effective domain generalization technique for histology compared to color jitter. Additionally, using only the highly cellular regions does not improve results, perhaps because the surrounding cells also reorganize in response to the mutated cells. Furthermore, using feature extraction pipelines trained on other histology data need not give better results. Finally, Among weakly supervised learning techniques, we got the best results with Clustering-constrained Attention Multiple Instance Learning (CLAM)~\cite{b5}. Lastly, since the TCGA data was collected under controlled conditions, we tested our technique data collected during clinical practice in Tata Memorial Centre in India, and obtained an AUC of 0.783, which suggests that this technique can generalize well if one has access to data for training from the same hospital on which it will be used.

Models trained on TCGA dataset did not perform well on the TMC dataset right out of the box, which suggests that stronger domain generalization techniques are needed for clinically deployable models.

\begin{figure}[!]
\centering
\includegraphics[height=5.5cm,width=6.5cm]{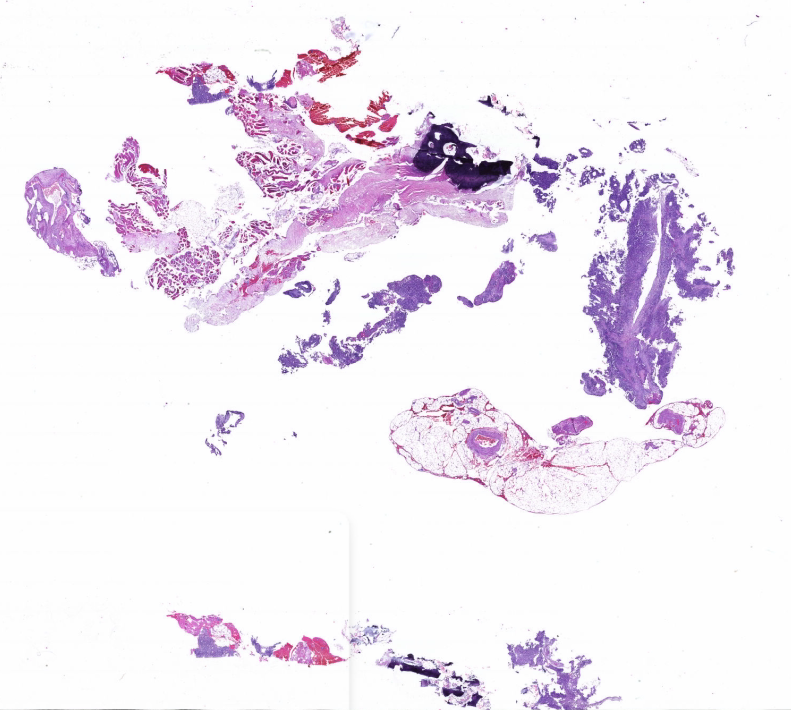}
\quad
\includegraphics[height=5.5cm,width=6.5cm]{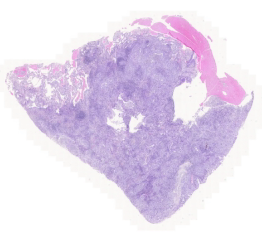}
\caption{Snapshot of one whole slide image each of H\&E stained FFPE lung cancer tissue from TMC (top) and TCGA (bottom) datasets}
\label{slide}
\end{figure}

\section{Related Work}

Obtaining pixel-level annotations for medical images is very difficult; this drastically reduces number of available data instances. However, obtaining a label for the entire image is easier by mining medical records. Therefore, it is appealing to divide a medical image into smaller patches, collectively considered as a bag with a single label \cite{b16}. This idea has attracted a great interest in the computational pathology. However, this approach leads to label noise where some patches marked with the disease class label may actually be a healthy or uninvolved tissue in the same WSI that contains diseased tissue.

Weakly supervised learning has shown to be useful in annotation-free training of deep learning models on WSIs. Multiple instance learning (MIL) is a form of weakly supervised learning where instance are arranged in bag and levels are provided for entire bag. Typically, most MIL approaches use max pooling or mean pooling~\cite{b8,b9,b10}. Both of these operations are non-trainable,  which limits the their applicability. In the classical work on MIL it is assumed that instances are represented by features that can be obtain using pre-trained networks~\cite{b3,b4,b7,b22}. \cite{b24} used a weakly supervised learning technique to train a DNN to predict BRAF V600E mutational status, determined using DNA testing, in H\&E stained images of thyroid cancer tissue without regional annotations. Recent work utilizes fully-connected neural networks (NN) in MIL and shows that it could still be beneficial \cite{b11}. For instance, \cite{b15} proposed attention based MIL but attention weights were trained as parameters of an auxiliary linear regression model. This form of MIL seems to particularly suitable for medical imaging where processing a WSI consisting of billions of pixels is a bottleneck for computation. Other noteworthy MIL approaches that have been used for histopathology data include Gaussian processes \cite{b17} and a two-stage approach with neural networks with an EM algorithm to determine instance classes \cite{b18}. Additionally, attention MIL with clustering \cite{b5} framework has been proposed for multi-class classification.

\begin{figure*}[!] 
\centering
% \RawFloats
% \begin{floatrow}
% \subfloat{
\includegraphics[height=4.5cm,width=15cm]{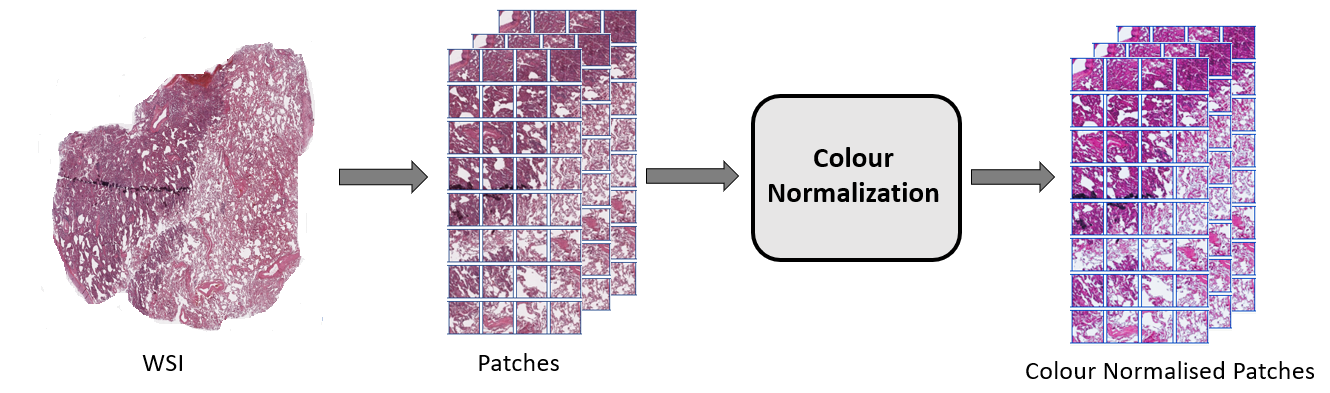}%}
\caption{A color normalization on representative WSIs patches of H\&E stained WSIs}
\label{fig2}
\vspace{5mm}
\subfloat{\includegraphics[height=4cm,width=14cm]{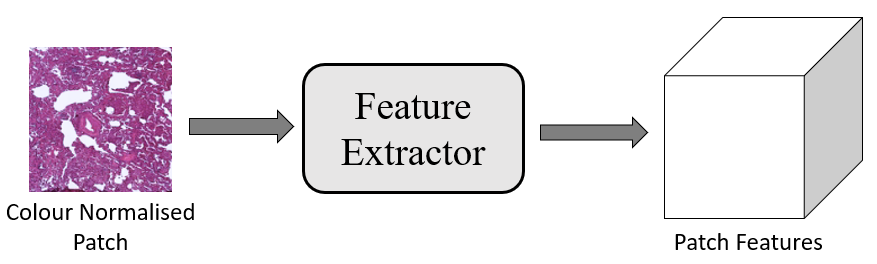}}
\caption{A pre-trained CNN extracts descriptive features from patches of H\&E stained WSIs}
\label{fig3}
% \end{floatrow}
\vspace{5mm}
\subfloat{\includegraphics[height=6cm,width=14.5cm]{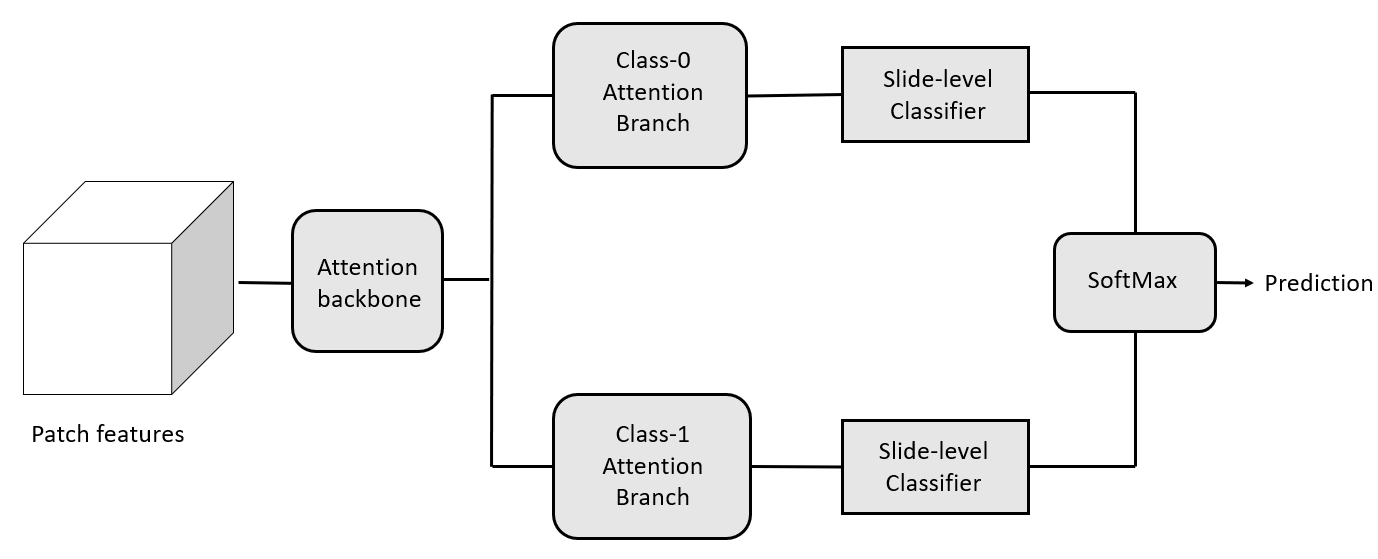}}
\caption{Feature vectors are fed to the model where an attention network aggregates patch-level information into slide-level representation, which are used to make final prediction}
\label{fig4}
\end{figure*}

\section{Proposed Method}

Patch extraction is one of the first steps while dealing with histopathology images. We extracted patches of size 512x512 pixels at 40x zoom level using OpenSlide library. The slides with a low amount of information were removed; i.e., all the patches (tiles) where greater than 50\% of the surface was covered by background, for which all the values are above 220 in the RGB color. 

Variations in staining protocols, reagent brands, habits of technicians, and scanner color response lead to color variation in digital histopathology images, which degrades the performance of deep learning models drastically. Therefore, the extracted patches were color normalized \cite{b25, b26}. 

From the color normalized patches we extracted features using ResNet50 trained on ImageNet \cite{b4}.

Weakly supervised classification task for pathology often involves a training set with known labels for each WSI, but no class-specific information or annotation is available for any pixel or region. Attention-based MIL~\cite{b1} with clustering builds on the MIL framework~\cite{b6} that is suitable for multi-class classification~\cite{b5}. MIL uses non-trainable aggregation function of max pooling, in which slide level prediction is based on the patch with the highest prediction probability, while attention MIL with clustering uses trainable and interpretable attention-based pooling function to aggregate slide level representation from patch level representation. In attention-based pooling, the attention network predicts two distinct sets of attention scores corresponding to the binary classification problem (in our case, EGFR versus non-EGFR). Because of this, our CNN learns which morphological features should be considered as positive evidence versus negative evidence for each class and computes two unique slide-level representations. We implemented CLAM \cite{b5} as a weakly supervised learning technique of choice (described below) but changed some of default hyperparameter settings based on our experiments. For a particular WSI represented as a bag of P instances or patches, we denote instance level embedding for $p^{\text{th}}$ patch using $e_p\in\mathbb{R}^{1024}$. After that $e_p$ is further compressed to 512-dimensional vector $k_p = W_1 e_p$  using the first fully connected layer $W_1\in R^{\text{512x1024}}$. Considering the first two layers $A_a\in R^{\text{256x512}}$ and $B_a\in R^{\text{256x512}}$ of the attention network (stacked fully connected layer) along with $W_1$ as a part of attention backbone shared by both the classes, this attention network splits into two parallel attention branches $W_{\text{a,1}}, W_{\text{a,2}} \in R^{\text{1x256}}$. To score class specific slide level representation two parallel independent classifiers ($W_{\text{c,1}}, W_{\text{c,2}}$) are trained. Attention score of $p^{\text{th}}$ patch for the $i^{\text{th}}$ class, denoted by $a_{\text{i,p}}$ is given by equation \ref{eq1} and slide-level representation aggregated per the attention score distribution for the $i^{\text{th}}$ class, denoted $k_{\text{slide,i}}\in R^{\text{512}}$, is given by equation \ref{eq2},

\begin{equation}
a_{\text{i,p}} =\frac{\exp{\{W_{\text{a,i}}(\tanh{(B_a k_p)}\odot sigm(A_a k_p))}\}}{\displaystyle \sum_{j=1}^{P} \exp{\{W_{\text{a,i}}(\tanh{(B_a k_j)}\odot sigm(A_a k_j))}\}} \label{eq1}
\end{equation}

\begin{equation}
k_{\text{slide,i}}=  \displaystyle\sum_{p=1} ^{P} a_{\text{i,p}}k_p\label{eq2}
\end{equation}

The slide level score is $s_{\text{slide,i}}$ = $W_{\text{c,i}}k_{\text{slide,i}}$ \cite{b5}. For regularization dropout is used after every layer in the attention backbone.

To improve the learning of class-specific features, binary clustering is used \cite{b5}. For each of the two classes, we planted a fully connected layer after the first layer $W_1$. The cluster assignment scores prediction for the $p^{\text{th}}$ patch as given by equation \ref{eq3}:

\begin{equation}
q_{\text{ i, p}} = W_{\text{inst,i}}k_{\text{p}}\label{eq3}
\end{equation}

Since we do not have patch-level labels the output of the attention network produces pseudo labels to supervise the clustering with the help of high and low attention scores. Therefore clustering is done by constraining patch level feature space $k_p$ in such a way that there is a linear separation between strong characterizing evidence and negative evidence for each class \cite{b5}. For instance level clustering smooth support vector machine (SVM) loss (based on multi-class SVM) is used. If the difference between the prediction score for the ground truth class and the maximum prediction score for the remaining class is greater than the specified margin, SVM loss penalizes the classifier linearly to the difference \cite{b5}.

During training a randomly sampled slide is provided to the model. To mitigate the class imbalance while training sampling probability of each slide is inversely proportional to the frequency of its ground truth class. Total loss for a slide $L_{\text{sum}}$ is composed of two-loss functions: (1) slide level classification loss $L_{\text{slide}}$ and (2) instance level clustering loss  $L_{\text{instance}}$, as given by equation \ref{eq4}:
\begin{equation}
 L_{\text{sum}}= a L_{\text{slide}} +b L_{\text{instance}}\label{eq4}
\end{equation}
where $a$ and $b$ are scaling hyperparameters. Here, $L_{\text{slide}}$ is standard cross entropy loss that compares the prediction score of a slide with ground truth slide-level label and $L_{\text{instance}}$ is binary SVM loss that compares instance level clustering prediction scores for each sampled patches with their corresponding pseudo-cluster labels \cite{b5}. 

We have used the constraint $a+b=1$ with $c= 0.7$. Additionally, we have examined the performance of our training methods using a 10-fold testing set up. We report the mean $\pm$ standard deviation as well as the maximum value of the area under receiver operating characteristic curve (AUC) in our results. We used Adam optimizer to update the model parameters with learning rate $4\times10^{-4}$ and weight decay of $1\times10^{-5}$. In all experiments, the running average of the first and the second moment of the gradient are computed with default coefficient ($\beta_1 =0.9$ and $\beta_2= 0.999$) and for numerical stability $\epsilon$ is set to $1\times10^{-8}$. All models are trained for between 50 and 200 epochs with an early stop criterion. When validation loss does not decrease from the previous low value over 20 consecutive epochs, the early stopping criterion is met and the model for the epoch with the best validation loss is used for evaluation on the test data.

\begin{table*}[h]
\begin{center}
\caption{Results from using ResNet50 as a feature extractor for tumor detection and histological subtyping}\label{t1}
\begin{tabular}{|p{3cm}|p{1.5cm}|p{1.5cm}|p{3cm}|p{2cm}|}
    \hline
    \textbf{Task} & \textbf{Trained on} & \textbf{Tested on} & \textbf{Avg. Test AUC $\pm$ STD} & \textbf{Max Test AUC} \\ \hline \hline
     Tumor vs Non-Tumor  & TCGA   & TCGA  &\textbf{0.964$\pm$0.064}  & 0.985      \\ \hline
    
     LUAD vs LUSC  & TCGA        & TCGA   &\textbf{0.942$\pm$0.014}        & 0.971   \\ \hline
     
\end{tabular}

\end{center}
\end{table*}

\begin{table*}[h]
\begin{center}
\caption{Results from using ResNet50 as a feature extractor for detecting EGFR mutation with and without color normalization (CN)~\cite{b25,b26} or nuclei filtering (NF), and CLAM~\cite{b5} versus attention MIL~\cite{b1}}\label{t2}
\begin{tabular}{|p{4cm}|p{1.5cm}|p{1.5cm}|p{3cm}|p{2cm}|}
    \hline
    \textbf{Model} & \textbf{Trained on} & \textbf{Tested on} & \textbf{Avg. Test AUC $\pm$ STD} & \textbf{Max Test AUC} \\ \hline \hline
     CN + NF + Attention MIL  & TCGA   & TCGA  &\textbf{0.663$\pm$0.083}  & 0.792      \\ \hline
     CN + CLAM (recommended)  & TCGA   & TCGA   &\textbf{0.865$\pm$0.060}       & 0.955     \\ \hline
     Attention MIL  & TMC   & TMC  &\textbf{0.767$\pm$0.043}  & 0.836      \\ \hline
     CN + CLAM (recommended)  & TMC        & TMC         &\textbf{0.781$\pm$0.063}  & 0.924 \\ \hline
\end{tabular}

\end{center}
\end{table*}

\begin{table*}[h]
\begin{center}
\caption{Results from testing the proposed pipeline on different datasets}\label{t3}
\begin{tabular}{|p{3cm}|p{1.5cm}|p{1.5cm}|p{3cm}|p{2cm}|}
    \hline
    \textbf{Task} & \textbf{Trained on} & \textbf{Tested on} & \textbf{Avg. Test AUC $\pm$ STD} & \textbf{Max Test AUC} \\ \hline \hline
    
    EGFR vs Non-EGFR  & TCGA        & TMC         & $0.531\pm0.037$   & $0.588$ \\ \hline
    
    EGFR vs Non-EGFR  & TMC        & TCGA         & $0.583\pm 0.067$  & $0.631$ \\ \hline
\end{tabular}

\end{center}
    
\end{table*}

\begin{table*}[h]
\begin{center}
\caption{Results from testing different feature extractors}\label{t4}
\begin{tabular}{|p{2.70cm}|p{1cm}|p{2.30cm}|p{3cm}|p{2cm}|}
    \hline
    \textbf{Task} & \textbf{Dataset} & \textbf{Feature Extractor} & \textbf{Avg. Test AUC $\pm$ STD} & \textbf{Max Test AUC} \\ \hline \hline
     EGFR vs Non-EGFR  & TMC        & KimiaNet         & $0.759\pm0.037$  & $0.852$\\ \hline
     EGFR vs Non-EGFR  & TMC        & SimCLR         & $0.753\pm0.037$  &$0.853$\\ \hline
\end{tabular}

\end{center}
    
\end{table*}

\section{Experiments and Results}
\label{sec:expt}
We evaluated slide-level classification performance of our pipeline for three clinical diagnostic tasks: tumor versus non-tumor classification, LUAD versus LUSC histological sub-typing, and EGFR mutation detection in LUAD using 10-folds. That is, we divided the data into ten folds with roughly the same proportion for each class. In each of the ten rounds, one fold was used for testing while the other nine folds were randomly split into training and validation sets. During training, we created a batch of 512 patches sampled randomly from slide in the training set. To get slide-level prediction, the pipeline first makes patch-level predictions and then averages their probability score. We validated our model after every 100,000 patches with an early stopping criteria on model when validation loss does not decrease for 20 consecutive validation epochs. The model with the minimum validation loss was evaluated on the test set. We report the mean, standard deviation, and maximum of the area under receiver operating characteristic curve (AUC) for the ten test folds for all our experiments.

Table \ref{t1} summarizes the results of our approach for cancer detection. The TCGA dataset was split in the ratio of 80:10:10 for training, validation and testing, respectively, for our all tasks. We achieved average test AUC of $0.964\pm0.064$ for the task of tumor versus non-tumor classification for over 1500 patients' frozen slide available with labels by using cross entropy loss as a bag loss. Model parameters were optimized using the Adam optimizer with a learning rate of $5\times10^{-4}$ and weight decay of $1\times10^{-5}$, with $\beta_1 =0.9$ and $\beta_2= 0.999$ with $\epsilon$ value of $1\times10^{-8}$. 

We performed other experiments with the same experimental setup. For sub-typing of lung cancer, LUAD versus LUSC, we achieved average test AUC of $0.942\pm0.014$ over 1045 patients formalin-fixed paraffin embedded (ffpe) slides available with labels, as shown in Table~\ref{t1}.

Our results for EGFR mutation detection are summarized in Table~\ref{t2} for ffpe slides, where we examine the use of color nomralization, nuclei filtering, and two weakly supervised learning methods -- CLAM~\cite{b5} with attention-based multiple instance learning~\cite{b1}. For EGFR mutation prediction the average test AUC achieved is $0.865\pm0.060$ over $179$ patients slides whose labels were available. We trained our model for EGFR mutation prediction for Indian dataset from TMC and achieved $0.781\pm0.063$ average test AUC over 544 patients slides with the same value of parameters used for TCGA dataset.

To confirm our model's suitability for independent cohorts, we tested our model on the TMC dataset after training on the TCGA dataset, and found relatively poor performance as an average AUC of $0.531\pm0.037$. In a similar manner when we tested our model on the TCGA dataset after training on the Indian dataset, we obtained an average AUC of $0.583\pm0.067$ as summarized in the table \ref{t3}. These results and visual inspection point to significant visual differences between the two datasets. Besides having different stain colors for H\&E, higher cancer grade and significant tar deposits were much more frequent in lung tissue from India, indicating a stronger prevalence of smoking-related lung cancer. Additionally, TMC is known to have a skew towards late stage cancers as there are no screening programs for lung cancer in India as opposed to the US, and it is a hospital of last resort for a large section of the population.

We conducted a set of ablation studies to understand the impact of the other components of our model as well. Informative features are the key to weakly supervised method for classification. Therefore we trained our model with feature extractors other than ResNet50 trained using ImageNet \cite{b4}, such as those trained specifically on histology images using self-supervised contrastive learning SimCLR~\cite{b7} and KimiaNet~\cite{b3}. A summary of these results is shown in table \ref{t4} for TMC dataset. Features from KimiaNet were fed to the attention MIL with clustering, attaining an average test AUC of $0.759$. When SimCLR was used as an feature extractor we achieved average test AUC of $0.753$. These results show that using feature extractors trained on pathology data is not always advantageous, especially if the feature extractor is trained on a different magnification and for an easier task -- KimiaNet is trained on 20x for organ and histological subtype detection, while we were performing mutation detection at 40x.

\section{Conclusion and Discussion}
We demonstrated our pipeline's effectiveness for tumor biomarker discovery from WSIs. The tumor classification model in our pipeline predicts whether the examined tissue is tumorous or normal with very high accuracy in line with previous results. Further, the histological subtype detection model in our pipeline can differentiate lung cancer sub-types of the TCGA dataset with high accuracy as well, again in line with previous results. Although no morphological signal is directly visible to a pathologist for detecting EGFR mutation, we show that appropriate pipelines with color normalization and weakly supervised deep learning models can predict EGFR mutation with an encouraging AUC for both TCGA and the TMC datasets. For the TCGA dataset, we were able to outperform previous studies on EGFR detection~\cite{b2}. Our ablation study found that KimiaNet pre-trained feature extractor models do not outperform conventional ResNet50 models pre-trained on ImageNet. The observation remained unchanged for the SimCLR-based feature extractor model as well.

We also performed a few additional experiments. In one of our experiments, we explicitly filtered out patches with fewer nuclei from our framework to enhance the feature learning in our training scheme. However, the performance in these cases was worse compared to our reported results. Further, we also tried to apply our model, trained on TCGA, on the TMC dataset and vice versa for EGFR prediction. We observed substantial performance degradation in both cases, affirming strong distribution shift between the datasets. Between the two datasets, we noticed significant differences due to variations in tissue preparation and staining, tar deposits, and cancer stage. Although we employed color jitter and normalization methods to reduce the distribution differences between the datasets, the performance did not improve much. In the future, we would like to experiment with and develop domain adaptation and domain generalization techniques to counter this problem.

\end{document}